\documentclass[conference]{IEEEtran}
\IEEEoverridecommandlockouts
\usepackage{cite}
\usepackage{amsmath,amssymb,amsfonts}
\usepackage{graphicx}
\usepackage{float}
\usepackage{textcomp}
\usepackage{xcolor}
\usepackage{enumitem}
\usepackage{booktabs}
\usepackage[square,numbers]{natbib}
\graphicspath{ {./img/} }
\usepackage[linesnumbered,lined,boxed,ruled]{algorithm2e}
\usepackage{tikz}

\newcommand*\circled[1]{\tikz[baseline=(char.base)]{
            \node[shape=circle,color=black, draw,inner sep=0.8pt, minimum size=2pt] (char) {#1};}}
\newcommand{\rpoint}[1]{\circled{{\fontfamily{pcr}\selectfont\footnotesize{#1}}}}

\def\BibTeX{{\rm B\kern-.05em{\sc i\kern-.025em b}\kern-.08em
    T\kern-.1667em\lower.7ex\hbox{E}\kern-.125emX}}

\usepackage{fancyhdr,lipsum}
\fancypagestyle{firstpage}{
  \fancyhf{}
  \fancyhead[C]{To appear at the 2022 International Joint Conference on Neural Networks (IJCNN), \\
  at the  2022 IEEE World Congress on Computational Intelligence (WCCI), July 2022, Padua, Italy.}
  \fancyfoot[C]{\thepage}
}

\begin{document}

\title{fakeWeather: Adversarial Attacks for Deep Neural Networks Emulating Weather Conditions on the Camera Lens of Autonomous Systems
}


\author{\IEEEauthorblockN{Alberto Marchisio\textsuperscript{1,*}\thanks{*These authors contributed equally to this work.}, Giovanni Caramia\textsuperscript{2,*}, Maurizio Martina\textsuperscript{2}, Muhammad Shafique\textsuperscript{3}}
\IEEEauthorblockA{\textit{\textsuperscript{1}Technische Universit{\"a}t Wien, Vienna, Austria}\ \ \ \textit{\textsuperscript{2}Politecnico di Torino, Turin, Italy}\ \ \ \textit{\textsuperscript{3}New York University, Abu Dhabi, UAE}} 
\IEEEauthorblockA{\textit{Email: alberto.marchisio@tuwien.ac.at, giovanni.caramia@studenti.polito.it}}
\IEEEauthorblockA{\textit{maurizio.martina@polito.it, muhammad.shafique@nyu.edu}}\\
\vspace*{-30pt}}


\maketitle
\thispagestyle{firstpage}

\begin{abstract}
Recently, Deep Neural Networks (DNNs) have achieved remarkable performances in many applications, while several studies have enhanced their vulnerabilities to malicious attacks. In this paper, we emulate the effects of natural weather conditions to introduce plausible perturbations that mislead the DNNs. By observing the effects of such atmospheric perturbations on the camera lenses, we model the patterns to create different masks that fake the effects of rain, snow, and hail. Even though the perturbations introduced by our attacks are visible, their presence remains unnoticed due to their association with natural events, which can be especially catastrophic for fully-autonomous and unmanned vehicles. We test our proposed \textit{fakeWeather} attacks on multiple Convolutional Neural Network and Capsule Network models, and report noticeable accuracy drops in the presence of such adversarial perturbations. Our work introduces a new security threat for DNNs, which is especially severe for safety-critical applications and autonomous systems.
\end{abstract}

\begin{IEEEkeywords}
Deep Neural Networks, Adversarial Attacks, Weather, Rain, Snow, Hail.
\end{IEEEkeywords}

\section{Introduction}

In the last decade, Deep Neural Networks (DNNs) have lifted groundbreaking advancements in several fields, including object recognition~\cite{Jiao2019DLObjectDetectionSurvey}, autonomous systems~\cite{Kuutti2019SurveyDLAutonomousVehicle}, video, image, and signal processing~\cite{Royson2021SurveyDNNStreaming}, and achieving the human-level or even more classification accuracy for certain tasks~\cite{He2016ResNet}. However, despite their great success~\cite{Capra2020SurveyDNNAccess}\cite{Capra2020Updated}, DNNs have been proven to be vulnerable to adversarial attacks, which undermine their security since they maliciously subvert the DNN predictions~\cite{Biggio2018WildPatterns}. While several works of adversarial machine learning have been proposed earlier~\cite{Dalvi2004Adv1}\cite{Lowd2005Adv2}, their first application to DNN-based algorithms was conducted by Szegedy et al.~\cite{Szegedy2014Intriguing}, who demonstrated that DNNs can easily be fooled by injecting imperceptible perturbations into the input images.

The usage of DNN-based algorithms for safety-critical applications requires that the DNNs do not fail their predictions under challenging conditions~\cite{Shafique2021ICCAD21SS}\cite{Dave2022AgileMethodology}. Such situations can appear in many different forms, including process variations that induce hardware faults, input pollution, or poisoning that induce a misbehavior~\cite{Shafique2020RobustML}. For instance, for vision applications in smart transportation systems, the DNNs should correctly work under different lights and atmospheric phenomena. Hence, an image captured in such conditions represents a plausible image that can be processed by the DNN-based algorithm. 

\subsection{Target Research Problem and Associated Challenges}

The key objective for an adversarial attack and its applicability in practical use-cases consists of not being recognized as adversarial, but rather as common/plausible. The most intuitive approach is to inject a very limited amount of perturbations, with the goal of making the differences between the clean images and the adversarial images imperceptible to the human eye. This approach has been adopted by several works, including Luo et al.~\cite{Luo2018ImperceptibleRobust}, Croce et al.~\cite{Croce2019ImperceivableAttacks}, and Marchisio et al.~\cite{Marchisio2019CapsAttacks}. However, the attacker needs to have access to a set of information, including DNN model architecture and parameters, inputs, and outputs (i.e., in \textit{white-box} settings), or only inputs and outputs (in \textit{black-box} settings). Even the most advanced decision-based black-box attacks such as HopSkipJumpAttack~\cite{Chen2020HopSkipJumpAttack} and FaDeC~\cite{Khalid2020FaDeC} still have access to the DNN predicted output class for each image. However, in practical cases, it may be very complicated to obtain such information, due to the protection mechanisms applied by the DNN-based system developers~\cite{Xue2021DNNIPProtection}. Moreover, another key limitation resides in the fact that even the most \mbox{query-efficient} algorithms~\cite{Willmott2021YOQO} need to perform a certain number of queries (i.e., inference passes) to generate the adversarial perturbation, which may not be practical in case of stringent real-time constraints, because of the latency overhead caused by the queries.

\begin{figure}[h!]
	\centering
	\includegraphics[width=.9\columnwidth]{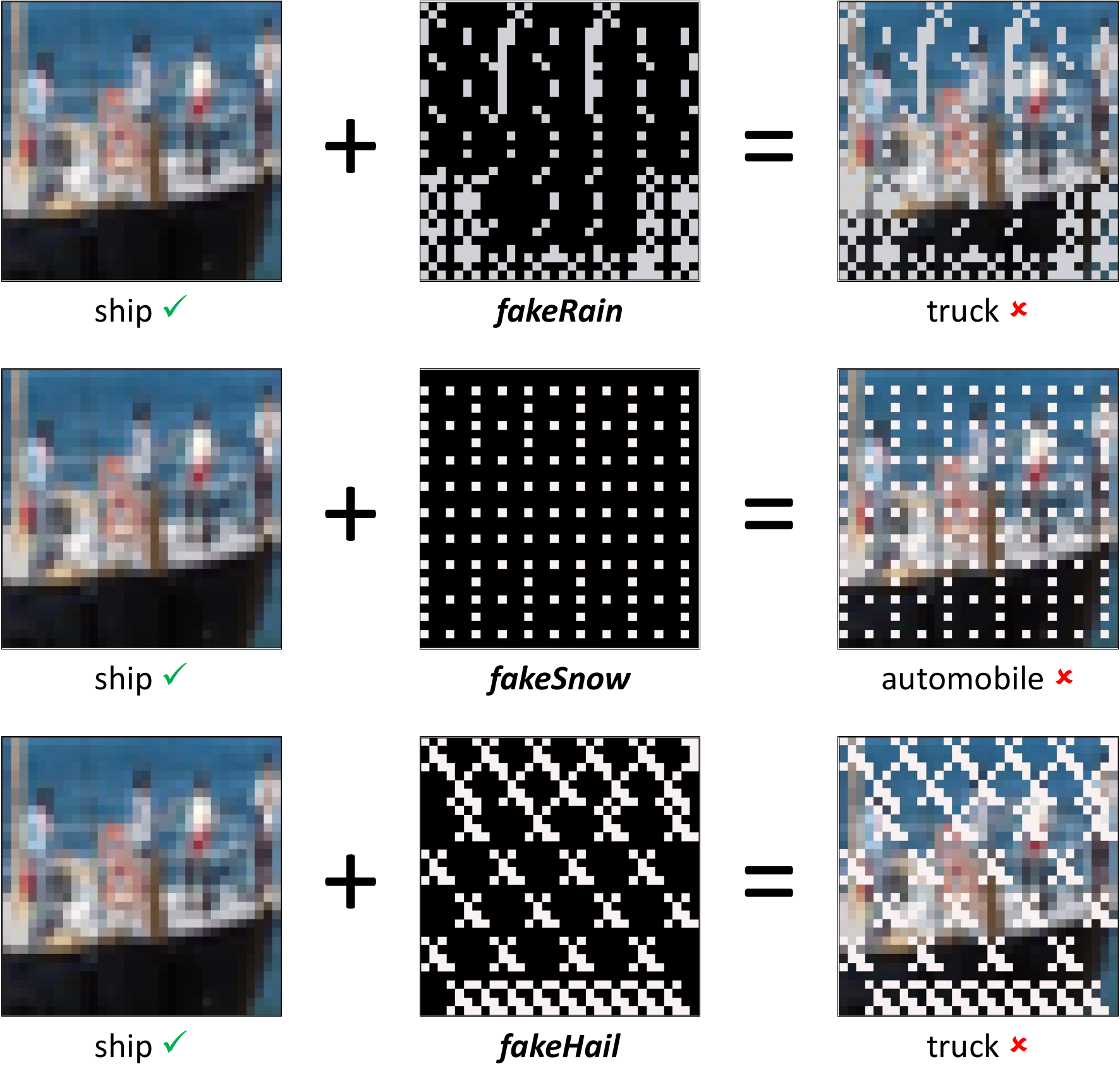}
	\caption{\textit{fakeWeather} attacks functionality.}
	\label{fig:fakeWeather_figure}
\end{figure}

Due to these limitations of the adversarial attacks that aim at introducing imperceptible perturbations compared to the original images, our approach follows a different strategy (see Figure~\ref{fig:fakeWeather_figure}). \textit{Our novel idea is to introduce perturbations to the input image in such a way that it is not considered as adversarial, since it resembles a natural condition captured by the camera.} While the differences between the clean image and the adversarial image can be noticed, the adversarial image itself is hardly categorized as ``adversarial'', since it simply captures a plausible natural condition. The reason is based on the fact that traditional adversarial machine learning takes into account the comparison between the adversarial image and the original image. However, in \mbox{real-world} practice, it is impossible to obtain such a comparison, since the only accessible image is the one recorded by the camera. Noticeably, our approach is advantageous compared to previous works, since it is conducted in what we call a \textit{true black-box setting}, i.e., in a scenario in which the attacker has no information about the DNN model architecture and parameters, nor its outputs. The only information required is the size of the input image, for generating an adversarial mask of that size. Moreover, our attack does not require any query, Thus it can easily be applied at run time.

\subsection{Our Novel Contributions}

Towards this, we observe how natural weather conditions, such as rain, snow, and hail, are perceived by the camera. We exploit this observation by designing \textit{fakeWeather} attack algorithms that emulate these effects on the camera lens. An overview of its functionality is shown in Figure~\ref{fig:novel_contributions}. Our methodology can be used not only as an adversarial attack to mislead the DNN, but also as a data augmentation approach for reinforcing the DNN training under these conditions. Our contributions can be summarized as follows:

\begin{figure}[t!]
	\centering
	\includegraphics[width=\linewidth]{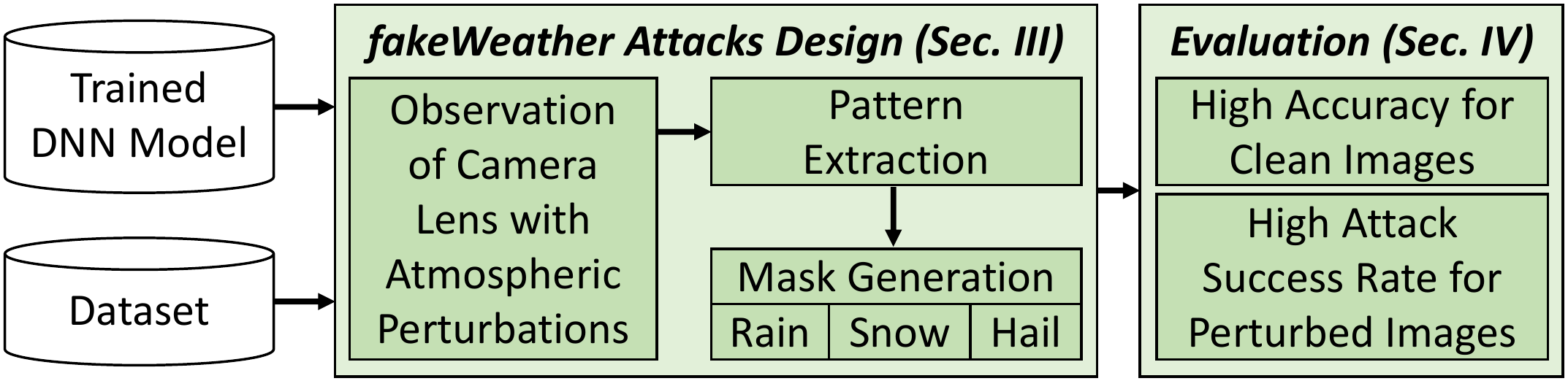}
	\caption{Overview of our Novel Contributions.}
	\label{fig:novel_contributions}
\end{figure}

\begin{itemize}
    \item We observe several images of natural weather events that affect the camera (\textbf{Section~\ref{subsec:weather_observation}}), and identify the patterns that are more commonly present in such images (\textbf{Section~\ref{subsec:pattern_extraction}}).
    \item By only knowing the image size, we design three \textit{fakeWeather} masks that fake the effect of such weather conditions on the camera lenses (\textbf{Sections~\ref{subsec:Rain_attack},~\ref{subsec:Snow_attack}, and~\ref{subsec:Hail_attack}}).
    \item We evaluate the \textit{fakeWeather} attacks on multiple DNN models (LeNet-5, ResNet-32, CapsNet) for the \mbox{CIFAR-10} dataset, and obtain a success rate of the attacks varying between $30\%$ and $82.5\%$ (\textbf{Section~\ref{sec:evaluation}}).
\end{itemize}

Before proceeding to the technical sections, we provide an overview of the adversarial attacks and the related works in Section~\ref{sec:background}.

\section{Background and Related Works}
\label{sec:background}

The purpose of the most common adversarial attack algorithms, such as gradient-based attacks~\cite{Goodfellow2015AdvExamples}, is to introduce some perturbations that induce a decision boundary cross in the DNNs, and therefore lead to a misclassification. Examples of such attacks include the Fast-Gradient Sign Method (FGSM)~\cite{Kurakin}, DeepFool~\cite{Moosavi-Dezfooli2016DeepFool}, the Projected Gradient Descent (PGD)~\cite{Madry2018PGD}, and the Carlini \& Wagner attack~\cite{CarliniWagner2016}. Other classes of attacks in which the perturbations were inserted only in a small set of pixels or only in one pixel were proposed by Narodytska et al.~\cite{Nina} and Su et al.~\cite{Su}, respectively. Concurrently, Moosavi-Dezfooli et al.~\cite{Moosavi-Dezfooli2017UniversalPerturbations} proposed image-agnostic universal perturbations that are applied to every sample, and Zhang et al.~\cite{Zhang2020CD-UAP} generated different adversarial perturbations for each target class.



In \textit{black-box} settings, several works were conducted. Kurakin et al.~\cite{Kurakin} proposed a method that crafts adversarial examples in the physical world by taking the images from a cell-phone camera. Moreover, taking into account the high-saliency and low-distortion path, Gragnaniello et al.~\cite{Gragn} introduced an attack that improves the perceptual quality of the adversarial image.


Several attacks have been designed for \textit{real-world} settings that incorporate so-called environmental perturbations. Brown et al.~\cite{Brown} generated adversarial patches, i.e., image-independent patches, to be placed anywhere inside the original image to mislead the DNNs. 
Following a similar approach, Eykholt et al.~\cite{Eykholt} added stickers to road signs to fool the traffic sign recognition system, while Sharif et al.~\cite{Sharif} added glasses to faces to fool the face recognition system. 
Man et al.~\cite{Man2020ghostimage} proposed GhostImage attacks, in which the adversarial patterns are inserted into the camera systems through a projector. 




Focusing on more closely related approaches to our work, other methods in which DNN models are fooled due to \textit{atmospheric phenomena} were proposed. Temel et al.~\cite{Temel2019CURE-TSD} analyzed several challenging conditions, including snow and rain, for traffic sign detection systems, and collected them into their proposed CURE-TSD-Real dataset. 
Zhai et al.~\cite{Zhai} simulated various rainy situations using a gradient-based rain generation process. 
However, there are clear differences compared to our \textit{fakeWeather} attacks. \textit{Both these two related works inject perturbations in the long-range, i.e., relatively far from the camera, while our methodology introduces perturbations in the close proximity of the camera lens. Unlike other methods in the related works, our approach is \mbox{non-invasive}, since it does not require any modification in the real world, but it only modifies some pixel intensities of the images without interfering with the underlying DNN processing.}



\section{fakeWeather Attack Design}
\label{sec:methodology}


\subsection{Problem Formulation and Assumptions}

\begin{figure}[h!]
	\centering
	\includegraphics[width=\columnwidth]{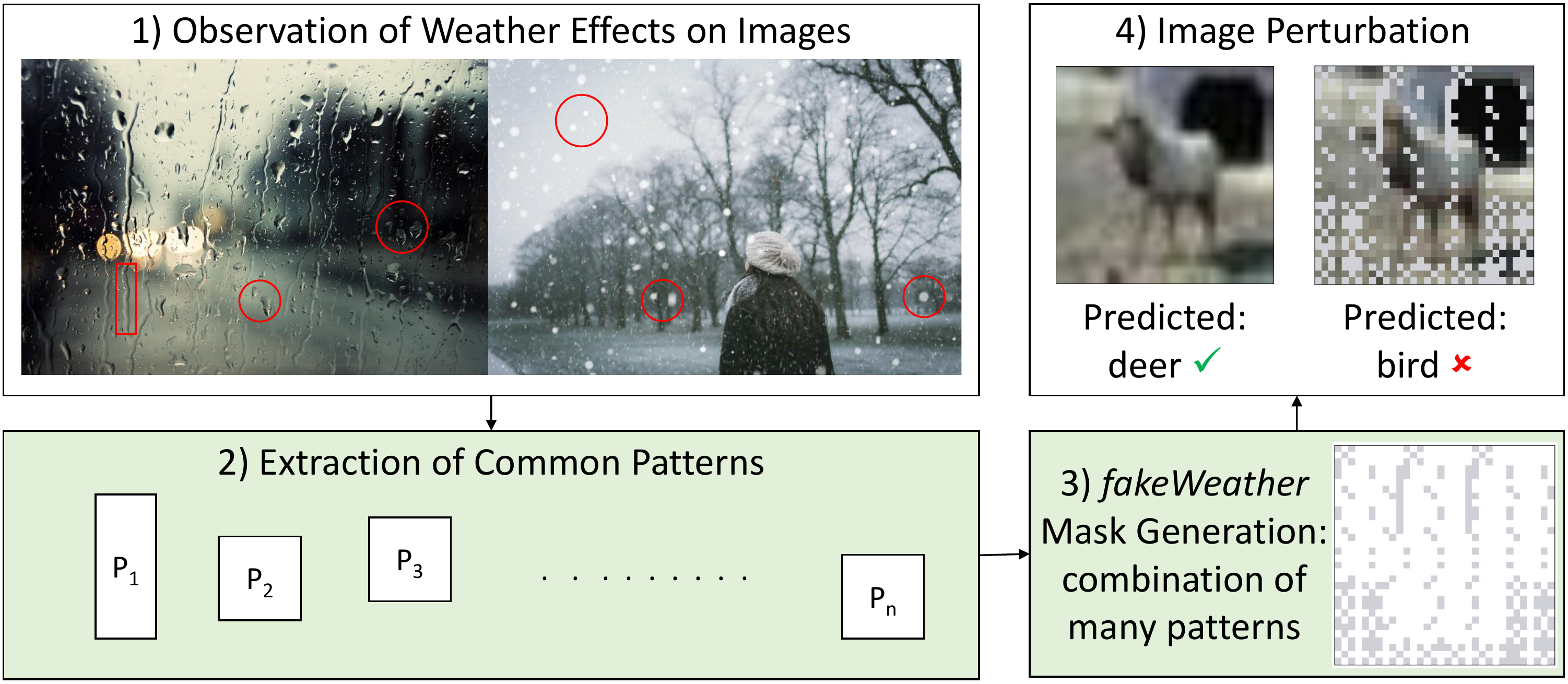}
	\caption{Overview of the \textit{fakeWeather} attack methodology.}
	\label{fig:methodology}
\end{figure}

Taking into account the previous discussions, we propose the \textbf{\textit{fakeWeather}} methodology. An overview of its functionality is shown in Figure~\ref{fig:methodology}. The final goal is to generate a finite set of perturbations with certain patterns which resemble the effect of natural weather events. Hence, such patterns are crafted by faking that the camera lens is dirty due to atmospheric conditions (such as rain, snow, and hail). After observing their effects on several examples in the real world, the common patterns are extracted and reproduced to generate the perturbation masks. The attack is conducted in what we call a \textit{true black-box setting}, i.e., assuming that:

\begin{itemize}
    \item the adversary has no information about the DNN model architecture, its parameters, and its output;
    \item the only information available for the attacker is the size of the input images.
\end{itemize}

\subsection{Observation of Weather Conditions}
\label{subsec:weather_observation}


The \textit{fakeWeather} attacks are performed through the introduction of drops of water and snowflakes. A typical water drop has a spherical shape, while a snowflake has a hexagonal shape. However, in practical use-cases, these weather conditions do not represent the main focus of the camera. A camera captures the effects of rain and snow in a different way, which results into a set of blurry dots that are overlapped to the image. For instance, if we consider the use case of vision for smart mobility, the camera can be placed either outside of the vehicle (and hence exposed to the weather conditions), or inside the vehicle but in close proximity to the window. Without loss of generality, we can model a drop or a snowflake as a single pixel w.r.t. the image of $h \cdot l$ pixels, where $h$ and $l$ represent the height and length, respectively.


\subsection{Pattern Extraction and Mask Generation}
\label{subsec:pattern_extraction}

According to the previous considerations, the \textit{fakeWeather} methodology extends the formulation of the One Pixel Attack~\cite{Su}, in which the perturbation of a \textit{single pixel} is defined as a tuple of 5 elements $(x, y, r, g, b)$ where:

\begin{itemize}
    \item  \textbf{$(x, y)$} represent the coordinates of the pixel to be modified;
    \item  \textbf{$(r, g, b)$} indicate the color of the pixel in RGB format.
\end{itemize}

\begin{figure}[h!]
	\centering
	\includegraphics[width=\columnwidth]{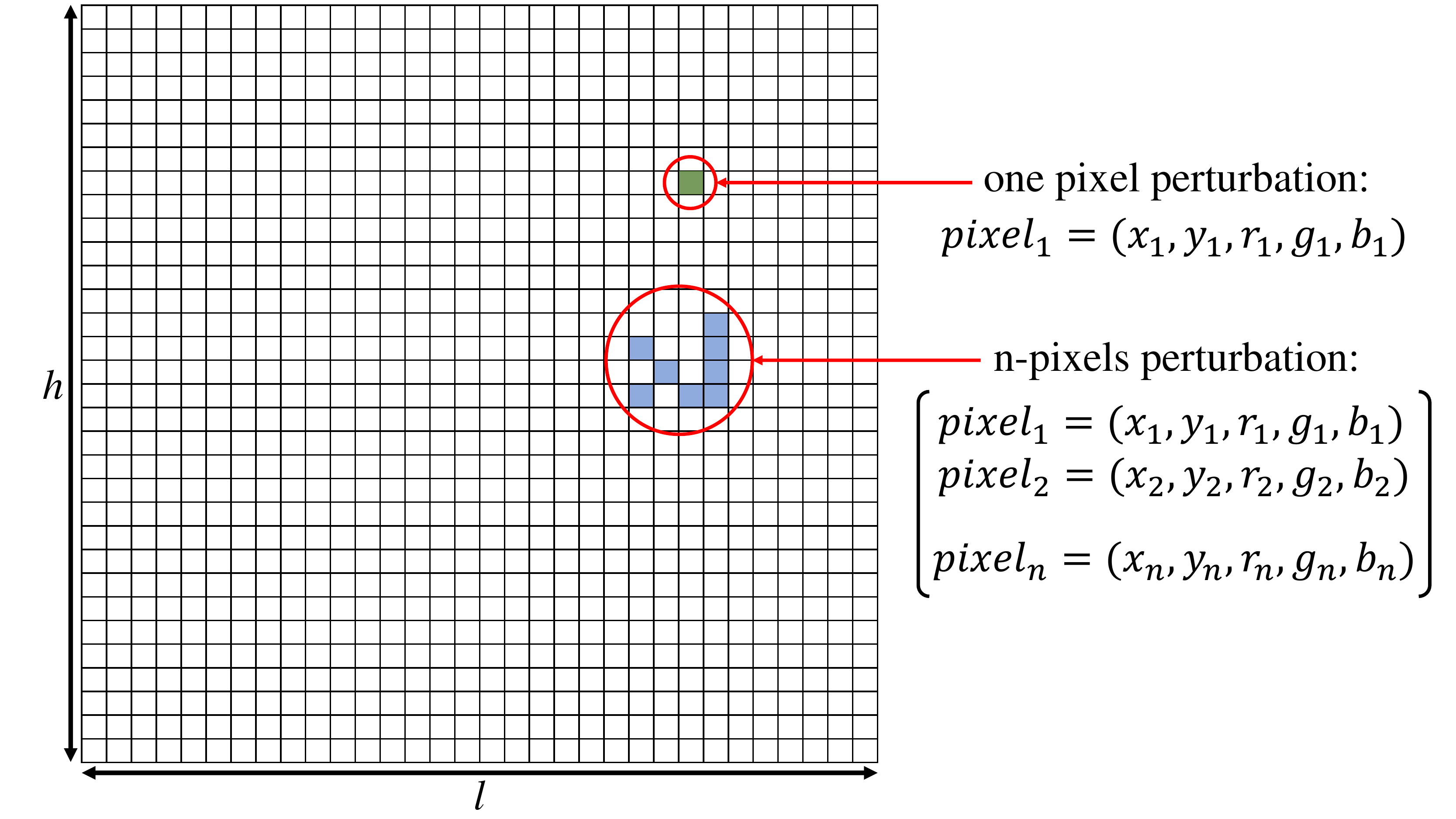}
	\caption{Encoding of pixel perturbations that form the adversarial pattern.}
	\label{fig:pattern_generation}
\end{figure}

Therefore, an adversarial pattern combines multiple pixel attacks, in which the perturbation introduced on the pixel $i$ can be written as in Equation~\ref{eq:pixel_perturbation}. An example of the corresponding mask of an adversarial pattern is shown in Figure~\ref{fig:pattern_generation}.

\begin{equation}
    pixel_i = (x_i, y_i, r_i, g_i, b_i)
    \label{eq:pixel_perturbation}
\end{equation}

The colors, i.e., the values assumed by $(r_i, g_i, b_i)$, are determined according to the weather condition:

\begin{itemize}
    \item \textbf{rain}: $(r_r, g_r, b_r) = (208, 209, 214)$
    \item \textbf{snow} and \textbf{hail}: $(r_s, g_s, b_s) = (249, 242, 242)$
\end{itemize}

For each type of \textit{fakeWeather} attack (i.e., rain, snow, and hail), specific patterns are generated. Common patterns are extracted from real images and reproduced to form the set of pixel coordinates $(x_i, y_i)$ that belongs to the attack mask. Once generated, the same mask is applied to all the images under attack.

\subsection{fakeRain Attack}
\label{subsec:Rain_attack}

\begin{figure*}[h!]
	\centering
	\includegraphics[width=\linewidth]{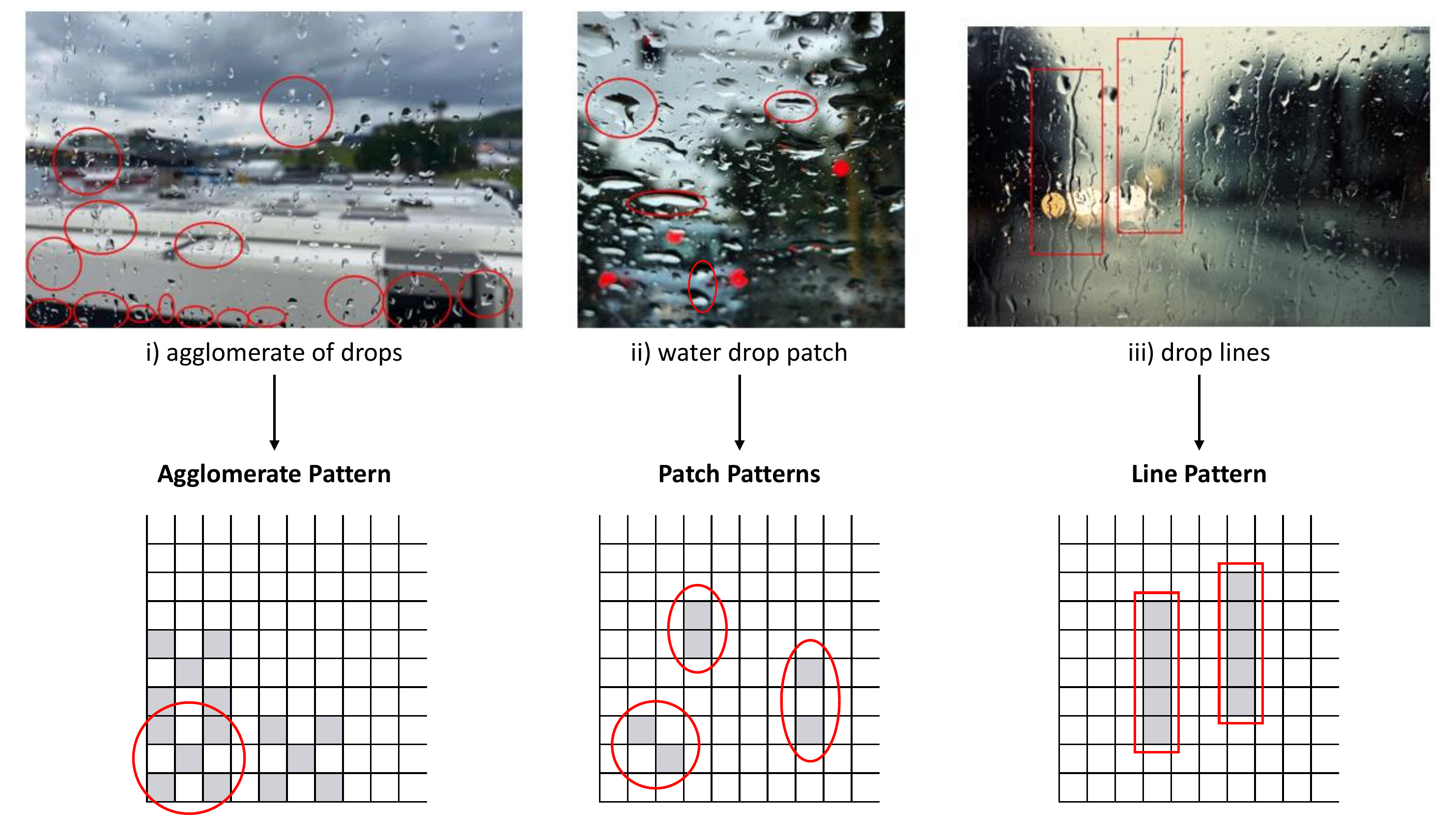}
	\caption{Several patterns of water drops observed from the real environment. i) agglomerate of drops, ii) water drop patch, iii) drop lines.} 
	\label{fig6}
\end{figure*}

The mask employed in the \textit{fakeRain} attack is designed based on the combination of several water drops. In the real world, the camera lens can be soiled due to the rain, where the water droplets make up different patterns. It is possible to recognize three real-case scenarios, which can be categorized as agglomerate of drops, drop patch, and drop lines. As shown in Figure~\ref{fig6}, the next step consists of modeling these patterns in terms of pixel coordinates that are perturbed.

\begin{figure}[h!]
\vspace*{-10pt}
\begin{minipage}[t]{.7\linewidth}
\begin{figure}[H]
	\centering
	\includegraphics[width=\linewidth]{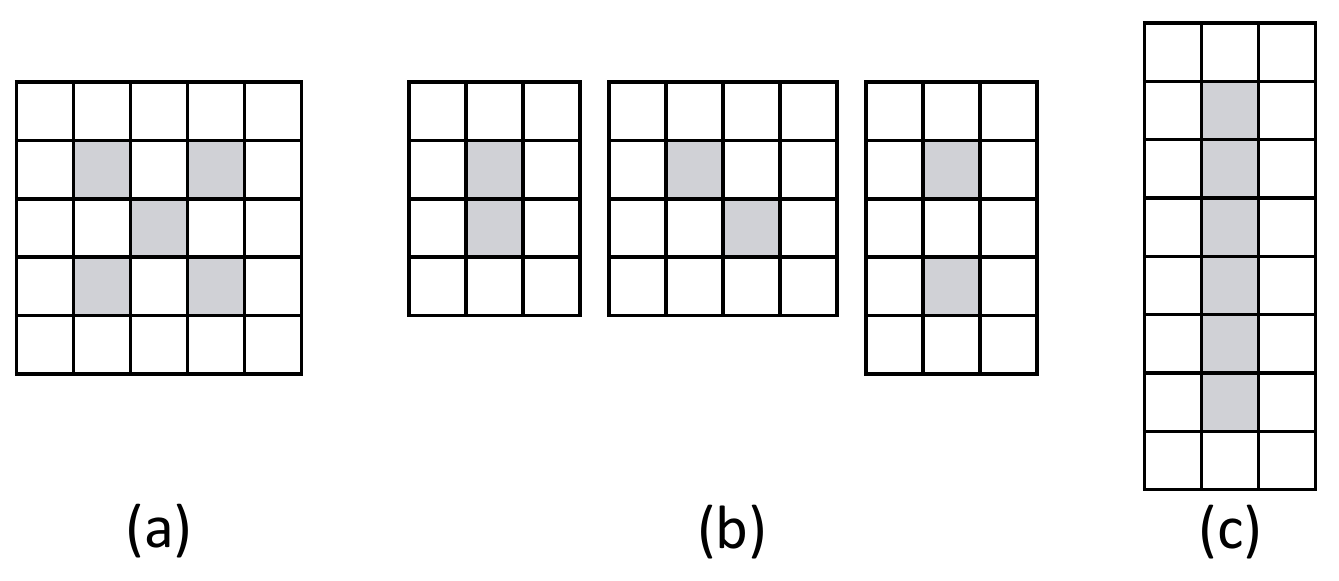}
	\caption{A graphical representation of (a) Agglomerate Pattern, (b) Patch Patterns, and (c) Line Pattern.}
	\label{fig:rain_patterns}
\end{figure}
\end{minipage}
\hfill
\begin{minipage}[t]{.26\linewidth}
\begin{figure}[H]
	\centering
	\includegraphics[width=\linewidth]{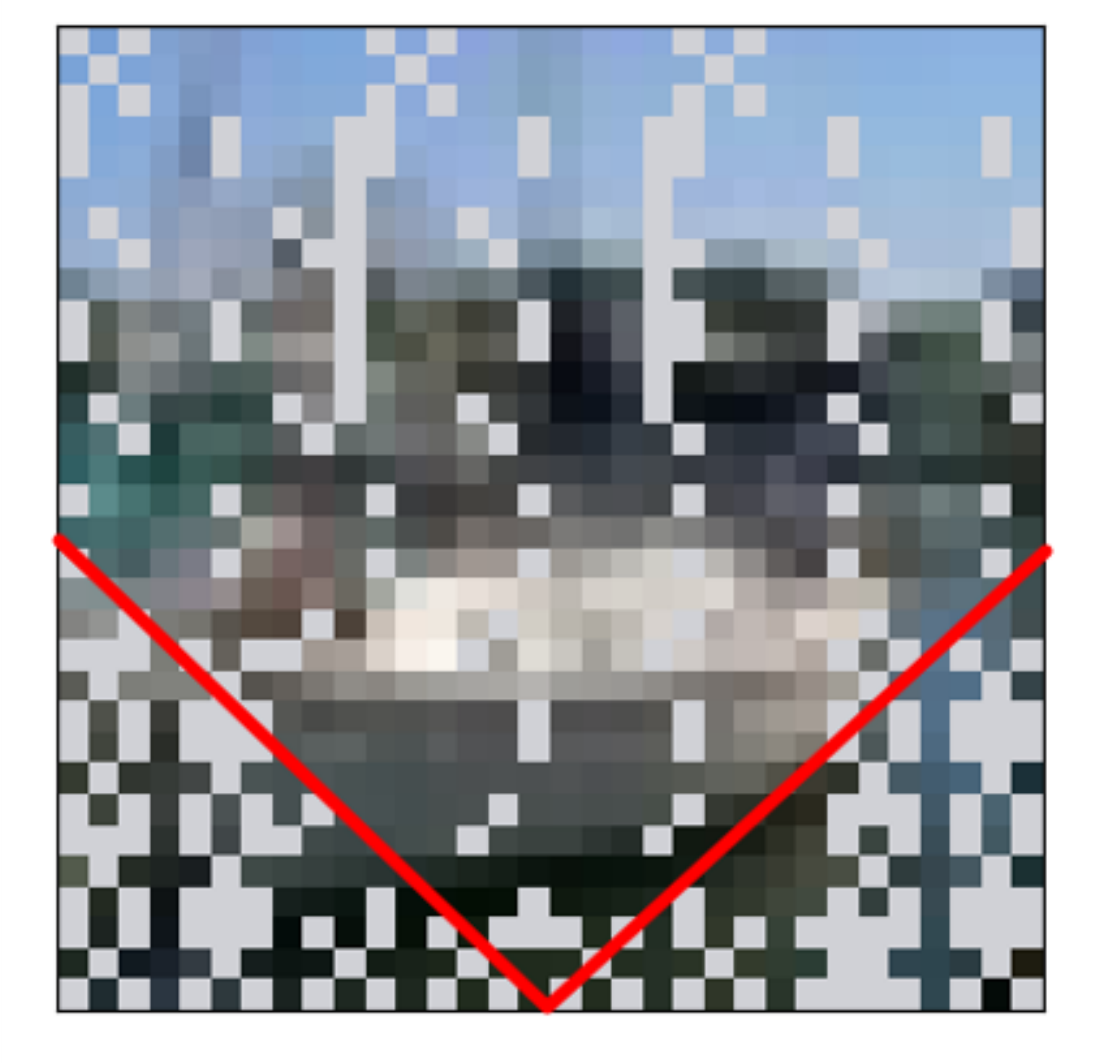}
	\caption{V-shaped fakeRain attack.}
	\label{fig9}
\end{figure}
\end{minipage}
\end{figure}

The \textit{Agglomerate Pattern} can be modeled by combining together 5 pixels to form a cross sign, according to the sketch in Figure~\ref{fig:rain_patterns}a and Algorithm~\ref{alg:Agglomerate_pattern}. The \textit{Patch Pattern} (see Figure~\ref{fig:rain_patterns}b) can have three different shapes, namely the \textit{vertical patch}, which can be modeled as two consecutive pixels that share the same $x$ coordinate (lines~4-6 of Algorithm~\ref{alg:patch_pattern}), the \textit{diagonal patch}, modeled as two pixels arranged to form a diagonal (lines~10-17 of Algorithm~\ref{alg:patch_pattern}, and the \textit{two dots patch}, in which two pixels are separated by a blank space (lines~20-22 of Algorithm~\ref{alg:patch_pattern}). The \textit{Line Pattern}, shown in Figure~\ref{fig:rain_patterns}c, is modeled as a vertical line of $n$ pixels (see Algorithm~\ref{alg:line_pattern}).

\IncMargin{1em}\begin{algorithm}
\SetKwData{Left}{left}
\SetKwData{This}{this}
\SetKwData{Up}{up}
\SetKwFunction{Union}{Union}
\SetKwFunction{FindCompress}{FindCompress}
\SetKwInOut{Input}{input}
\SetKwInOut{Output}{output}
\Input{Coordinate $(x_0,y_0)$}
\Output{Agglomerate Pattern $P_a$}
$P_a = \varnothing$ \\
$k=0$ \\
\For{$i \leftarrow 0$ \KwTo $2$}{
    \For{$j \leftarrow 0$ \KwTo $2$}{
        \If{$(i+j=0 \vee i+j=2 \vee i+j=4)$}{
            $P_a \leftarrow pixel_k = (x_0+i,y_0+j,r_r, g_r, b_r)$ \\
            $k \leftarrow k+1$ \\
        }
    }
}
\caption{Agglomerate Pattern}\label{alg:Agglomerate_pattern}
\end{algorithm}
\DecMargin{1em}




\IncMargin{1em}\begin{algorithm}
\SetKwData{Left}{left}
\SetKwData{This}{this}
\SetKwData{Up}{up}
\SetKwFunction{Union}{Union}
\SetKwFunction{FindCompress}{FindCompress}
\SetKwInOut{Input}{input}
\SetKwInOut{Output}{output}
\SetNoFillComment
\Input{Coordinate $(x_0,y_0)$, Type $t$}
\Output{Patch Pattern $P_p$}
$P_p = \varnothing$ \\
\Switch{$t$}{
    \Case(\tcp*[h]{Vertical Patch}){$0$}{
        \For{$j \leftarrow 0$ \KwTo $1$}{
            $P_p \leftarrow pixel_j = (x_0,y_0+j,r_r, g_r, b_r)$ \\
        }
    }
    \Case(\tcp*[h]{Diagonal Patch}){$1$}{
        $k=0$ \\
        \For{$i \leftarrow 0$ \KwTo $1$}{
            \For{$j \leftarrow 0$ \KwTo $1$}{
                \If{$(i+j=1)$}{
                    $P_p \leftarrow pixel_k = (x_0+i,y_0+j,r_r, g_r, b_r)$ \\
                    $k \leftarrow k+1$ \\
                }
            }
        }
    }
    \Case(\tcp*[h]{Two Dots Patch}){$2$}{
        \For{$j \leftarrow 0$ \KwTo $1$}{
            $P_p \leftarrow pixel_j = (x_0,y_0+2 \cdot j,r_r, g_r, b_r)$ \\
        }
    }
}

\caption{Patch Pattern}\label{alg:patch_pattern}
\end{algorithm}
\DecMargin{1em}

\IncMargin{1em}\begin{algorithm}
\SetKwData{Left}{left}
\SetKwData{This}{this}
\SetKwData{Up}{up}
\SetKwFunction{Union}{Union}
\SetKwFunction{FindCompress}{FindCompress}
\SetKwInOut{Input}{input}
\SetKwInOut{Output}{output}
\SetNoFillComment
\Input{Coordinate $(x_0,y_0)$, Length $n$}
\Output{Line Pattern $P_l$}
$P_l = \varnothing$ \\
\For{$j \leftarrow 0$ \KwTo $n-1$}{
    $P_l \leftarrow pixel_j = (x_0,y_0+j,r_r, g_r, b_r)$ \\
}
\caption{Line Pattern}\label{alg:line_pattern}
\end{algorithm}
\DecMargin{1em}


Moreover, in rainy conditions, we can notice that the water drops tend to concentrate in the bottom corners of the image. Hence, to emulate this effect, in the \textit{fakeRain} attack, a \textit{V-shape} is created to divide the image into two regions (see the example in Figure~\ref{fig9}). Below the \textit{V}, several agglomerate patterns are densely concentrated. Above the \textit{V}, path and line patterns are more sparsely distributed. Algorithm~\ref{alg:Rain_Attack} describes the procedure for generating the \textit{fakeRain} mask. Note that it is a three-step process in which (i) several agglomerate patterns are added (see line~2 of Algorithm~\ref{alg:Rain_Attack}), (ii) other agglomerate patterns are added if the coordinate is below the \textit{V} (line~5 of Algorithm~\ref{alg:Rain_Attack}), and (iii) patch patterns of different types and line patterns are added above the \textit{V} (line~7 of Algorithm~\ref{alg:Rain_Attack}).

\IncMargin{1em}\begin{algorithm}
\SetKwData{Left}{left}
\SetKwData{This}{this}
\SetKwData{Up}{up}
\SetKwFunction{Union}{Union}
\SetKwFunction{FindCompress}{FindCompress}
\SetKwInOut{Input}{input}
\SetKwInOut{Output}{output}
\Input{Image size: length $l$ and hight $h$}
\Output{fakeRain Mask $M_r$}
$M_r = \varnothing$ \\
$M_r \leftarrow P_a(\{0,...,l-3\},0)$ \\
\tcp{use many agglomerate patterns in the first line}
\For{$(i,j) \in (\{0,...,l-3\},\{0,...,h-3\})$}{
    \uIf{$(i+j<\frac{h+l}{4}) \vee (l-i+j<\frac{h+l}{4})$}{
        $M_r \leftarrow P_a(i,j) \vee \{\}$\\
        \tcp{sparsely add agglomerate patterns below the V}
    }
    \uElse{
        $M_r \leftarrow P_p(i,j,t) \vee P_l(i,j,n) \vee \{\}$\\
        \tcp{sparsely add patch patterns or line patterns above the V}
    }
}

\caption{fakeRain Mask Generation}\label{alg:Rain_Attack}
\end{algorithm}
\DecMargin{1em}


\subsection{fakeSnow Attack}
\label{subsec:Snow_attack}

\begin{figure}[h!]
	\centering
	\includegraphics[width=\columnwidth]{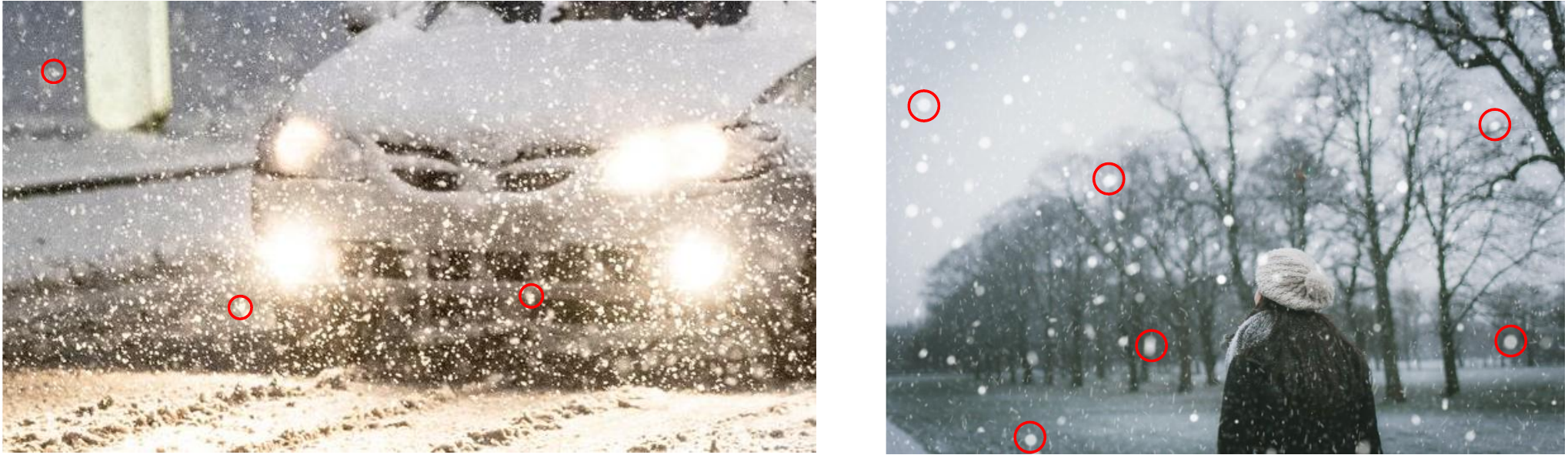}
	\caption{Several snowflakes observed, which can be modeled as single dots.}
	\label{fig:snow_observations}
\end{figure}

The design of the \textit{fakeSnow} attack is based on the assumption that a snowflake can be modeled as a single pixel, since the dimension of each snowflake is relatively small, as observed in Figure~\ref{fig:snow_observations}. According to these considerations, the snow pattern $P_s$ consists of a single pixel, which can be modeled as in Equation~\ref{eq:snow_pattern}, where $(x_0,y_0)$ represents the coordinate in which the snow pattern is constructed.

\begin{equation}
    P_s \leftarrow pixel_0(x_0, y_0, r_s, g_s, b_s)
    \label{eq:snow_pattern}
\end{equation}

\begin{figure}[h!]
	\centering
	\includegraphics[width=.5\linewidth]{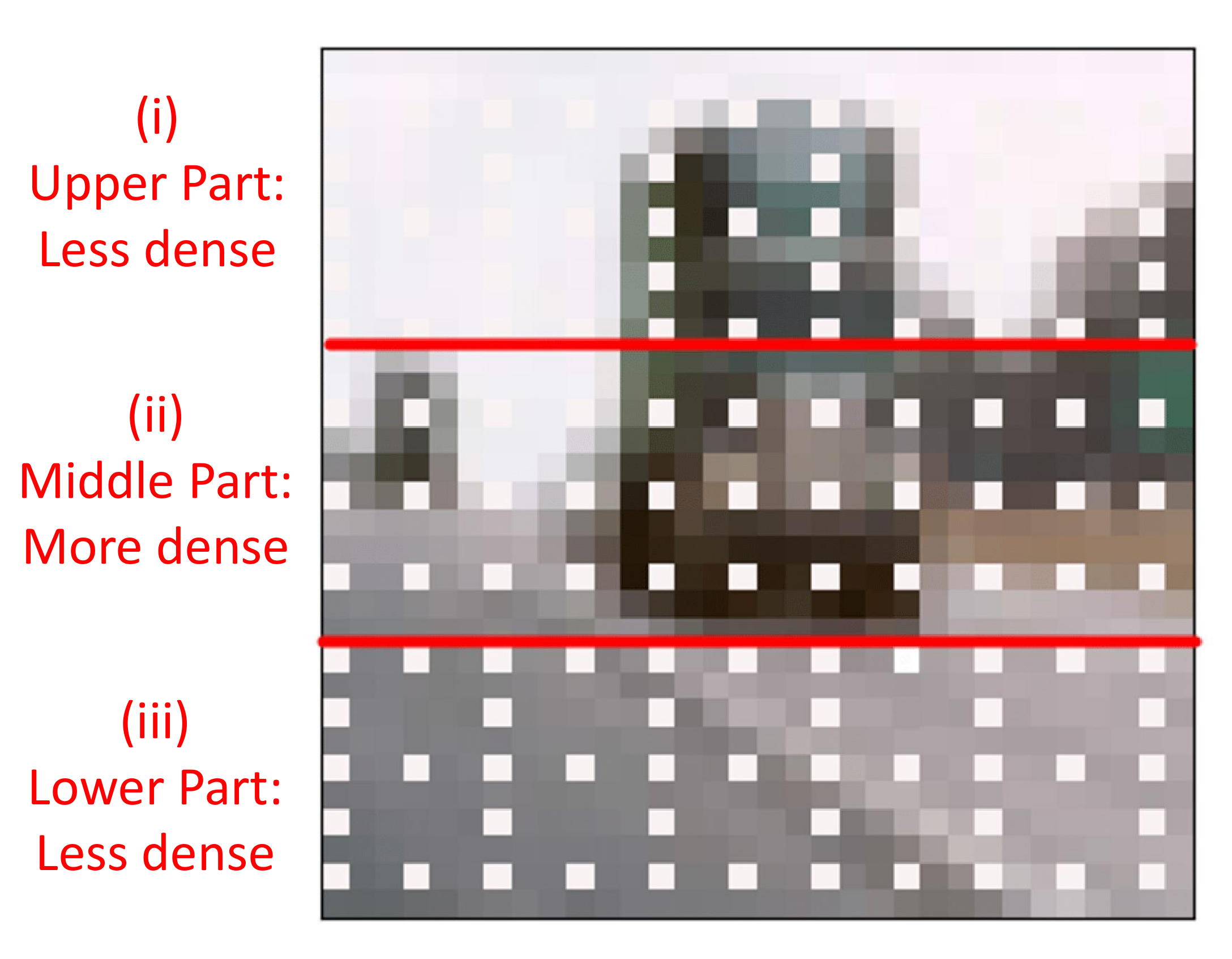}
	\caption{Mask for the fakeSnow attack, divided into three parts.}
	\label{fig:snow_3lines}
\end{figure}

Another key feature noticed from the observation of real images is that the snowflakes are more densely concentrated in close proximity to the horizon line. In practice, this effect can be modeled by cutting the image into three parts through two horizontal lines, as shown in Figure~\ref{fig:snow_3lines}, and placing more dense snow patterns in the middle region, while maintaining the top and the bottom of the image relatively less populated by snow patterns.
The generation of the mask for the \textit{fakeSnow} attack is described in Algorithm~\ref{alg:Snow_Attack}. It proceeds in different ways based on the vertical coordinate~$j$. In the middle part of the image, equally-spaced dense snow patterns are added to the \textit{fakeSnow} mask (line~9 of Algorithm~\ref{alg:Snow_Attack}). In the upper part and lower part of the image, alternating rows of dense and sparse (i.e., largely spaced) snow patterns are added (lines~4-7 of Algorithm~\ref{alg:Snow_Attack}).


\IncMargin{1em}\begin{algorithm}
\SetKwData{Left}{left}
\SetKwData{This}{this}
\SetKwData{Up}{up}
\SetKwFunction{Union}{Union}
\SetKwFunction{FindCompress}{FindCompress}
\SetKwInOut{Input}{input}
\SetKwInOut{Output}{output}
\Input{Image size: length $l$ and hight $h$}
\Output{fakeSnow Mask $M_s$}
$M_s = \varnothing$ \\
\For{$j, \in \{0,2,4,...,h-2\})$}{
    \uIf{$(j<\frac{h}{3}-1) \vee j>\frac{2h}{3}-1)$}{
        \tcp{upper and lower parts}
        \uIf{$j \equiv 0 \mod 4$}{
            $M_s \leftarrow P_s(\{0,3,6,9,...,l-2\},j+1)$\\
        }
        \uElse(\tcp*[h]{skip some snow patterns}){
            $M_s \leftarrow P_s(\{0,6,12,...,l-2\},j+1)$\\
        }
    }
    \uElse(\tcp*[h]{middle part}){
        $M_s \leftarrow P_s(\{0,3,6,9,...,l-2\},j+1)$\\
        \tcp{add dense snow patterns}
    }
}


\caption{fakeSnow Mask Generation}\label{alg:Snow_Attack}
\end{algorithm}
\DecMargin{1em}

\subsection{fakeHail Attack}
\label{subsec:Hail_attack}
 
\begin{figure}[h!]
	\centering
	\includegraphics[width=\linewidth]{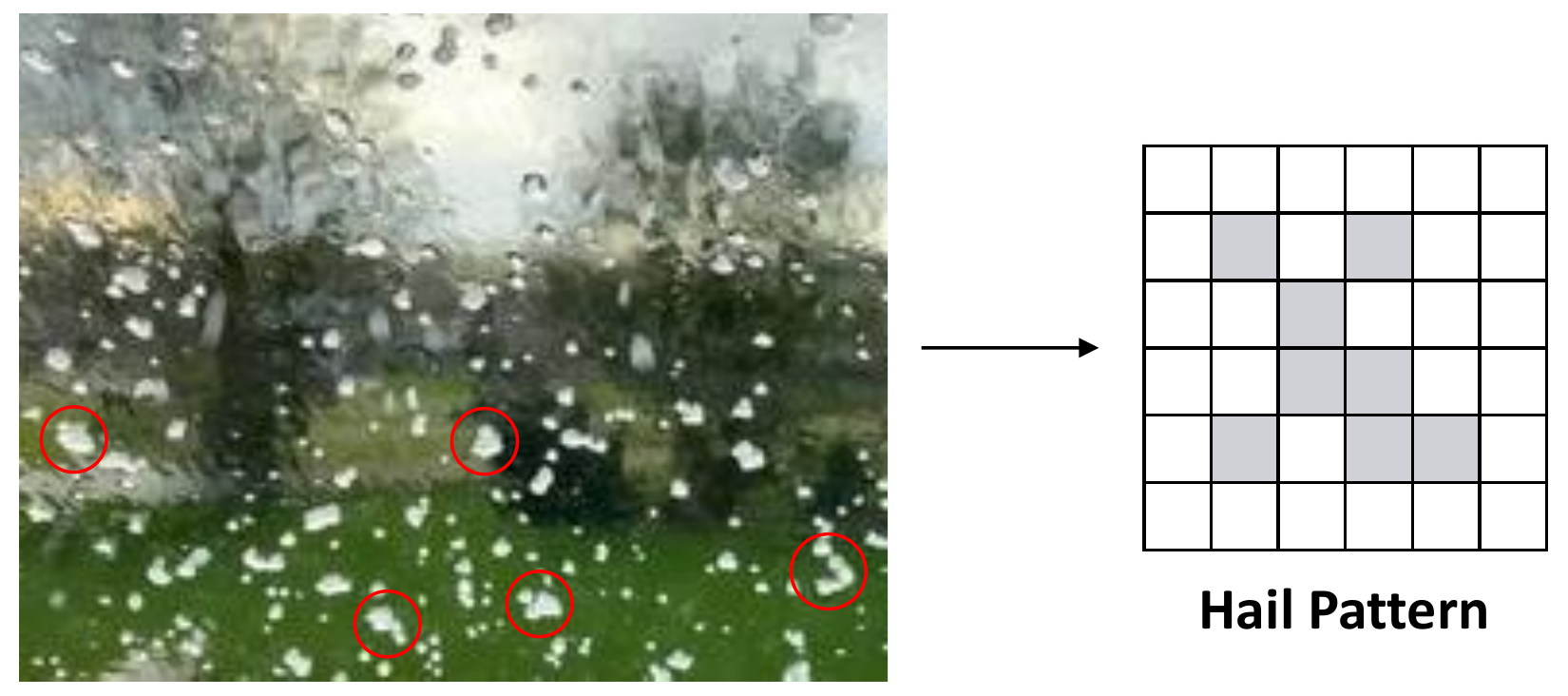}
	\caption{Observation of hail conditions, which lead to the design of the hail pattern.}
	\label{fig:hail_observations}
\end{figure}

Compared to the snow, a hail scenario produce relatively larger ice balls perceived by the camera, as shown in Figure~\ref{fig:hail_observations}. Hence, the hail pattern is not modeled as a single pixel, but as an agglomerate of 8 pixels, as described in Algorithm~\ref{alg:Hail_pattern}.

\IncMargin{1em}\begin{algorithm}
\SetKwData{Left}{left}
\SetKwData{This}{this}
\SetKwData{Up}{up}
\SetKwFunction{Union}{Union}
\SetKwFunction{FindCompress}{FindCompress}
\SetKwInOut{Input}{input}
\SetKwInOut{Output}{output}
\Input{Coordinate $(x_0,y_0)$}
\Output{Hail Pattern $P_h$}
$P_h = \varnothing$ \\
$k=0$ \\
\For{$i \leftarrow 0$ \KwTo $3$}{
    \For{$j \leftarrow 0$ \KwTo $3$}{
        \If{$(i=j \wedge i<2) \vee (i+j=3) \vee (i=2 \wedge j \neq 2)$}{
            $P_h \leftarrow pixel_k = (x_0+i,y_0+j,r_s, g_s, b_s)$ \\
            $k \leftarrow k+1$ \\
        }
    }
}
\caption{Hail Pattern}\label{alg:Hail_pattern}
\end{algorithm}
\DecMargin{1em}

Since the hail patterns appear irregularly, the \textit{fakeHail} mask can be generated through a collection of hail patterns, as described in Algorithm~\ref{alg:Hail_ Attack}. Note that the hail patterns are sparsely added, since, for each coordinate of the mask, the hail pattern can be added to the mask or not (line~3 of Algorithm~\ref{alg:Hail_ Attack}).

\IncMargin{1em}\begin{algorithm}
\SetKwData{Left}{left}
\SetKwData{This}{this}
\SetKwData{Up}{up}
\SetKwFunction{Union}{Union}
\SetKwFunction{FindCompress}{FindCompress}
\SetKwInOut{Input}{input}
\SetKwInOut{Output}{output}
\Input{Image size: length $l$ and hight $h$}
\Output{fakeHail Mask $M_h$}
$M_h = \varnothing$ \\
\For{$(i,j) \in (\{0,...,l-4\},\{0,...,h-4\})$}{
    $M_h \leftarrow P_h(i,j) \vee \{\}$\\
}

\caption{fakeHail Mask Generation}\label{alg:Hail_ Attack}
\end{algorithm}
\DecMargin{1em}



\section{Evaluating the Weather Attack}
\label{sec:evaluation}

\subsection{Experimental Setup}

\begin{figure}[h!]
	\centering
	\includegraphics[width=\linewidth]{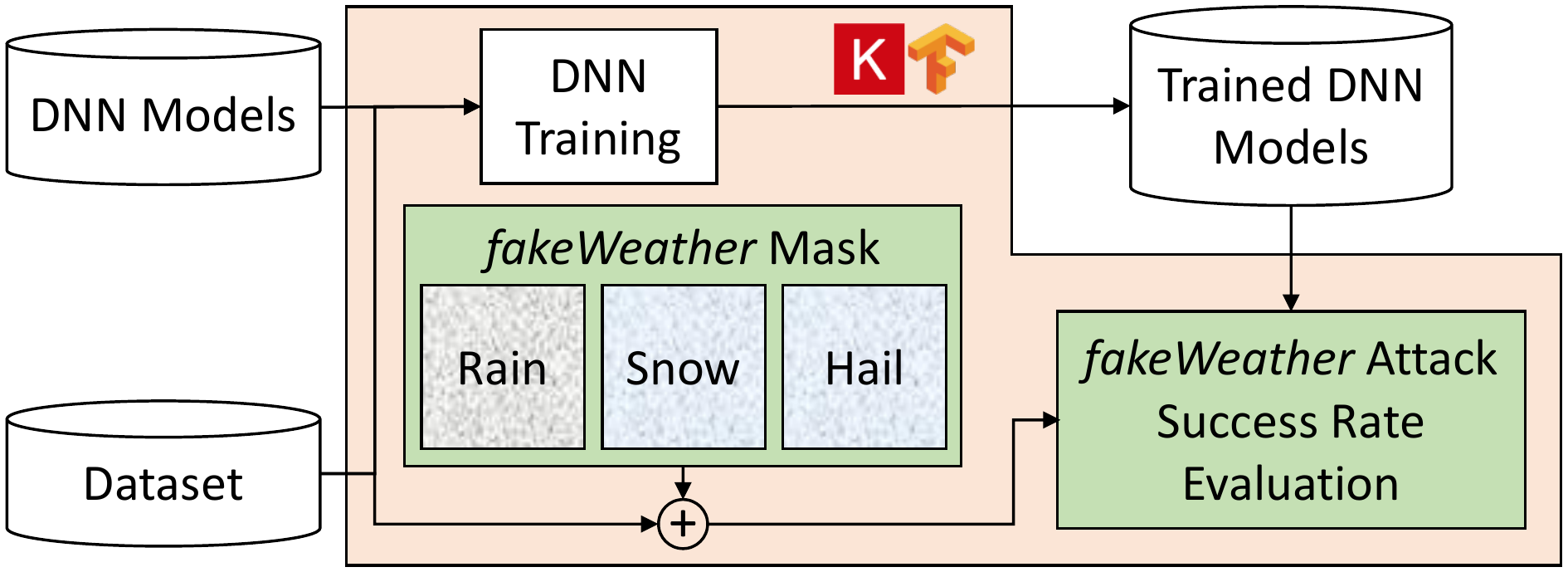}
	\caption{Experimental setup and tool-flow for conducting our experiments.}
	\label{fig:experimental_setup}
\end{figure}

We conducted the experiments on three different DNN models, which are the LeNet-5~\cite{Lecun1998LeNet}, the ResNet-32~\cite{He2016ResNet} and the CapsNet~\cite{Sabour2017CapsNet}, trained for the CIFAR-10 dataset~\cite{CIFAR-10}. It is a collection of $50,000$ training images and $10,000$ testing images of size $32 \times 32 \times 3$, divided into 10 classes. An overview of the setup and tool-flow employed for conducting the experiments is shown in Figure~\ref{fig:experimental_setup}.

The LeNet, which is composed of two convolutional layers and two fully-connected layers followed by a softmax layer, has been trained for $200$ epochs, using a batch size of $128$, weight decay $0.0001$, and a learning rate scheduler that progressively reduces its value from $0.05$ to $0.0004$. The \mbox{32-layer} ResNet has been trained for $200$ epochs, using a batch size of $128$, weight decay $0.0001$, and a learning rate scheduled to decrease from $0.1$ to $0.001$. The CapsNet, composed of a convolutional layer, a primary capsule layer, and a dynamic routing layer, has been trained for $200$ epochs with a batch size of $64$ and a learning rate equal to $0.001$. For clean test images, we measure the accuracy values of $74.88\%$, $92.31\%$, and $79.82\%$, for the LeNet, ResNet, and CapsNet, respectively.

Afterwards, the \textit{fakeWeather} masks have been applied to 200 testing samples and the attack success rate has been evaluated for every attack type (i.e., \textit{fakeRain}, \textit{fakeSnow} and \textit{fakeHail}) and every DNN model. The training, as well as the implementation of the \textit{fakeWeather} attacks and their evaluation, has been carried out using the Keras framework~\cite{chollet2015keras} with the TensorFlow~\cite{abadi2016tensorflow} back-end, and executed on an ML-workstation equipped with two Nvidia GeForce RTX 2080 Ti GPUs.

\subsection{fakeWeather Attacks Evaluation}

\begin{figure}[h!]
	\centering
	\includegraphics[width=\linewidth]{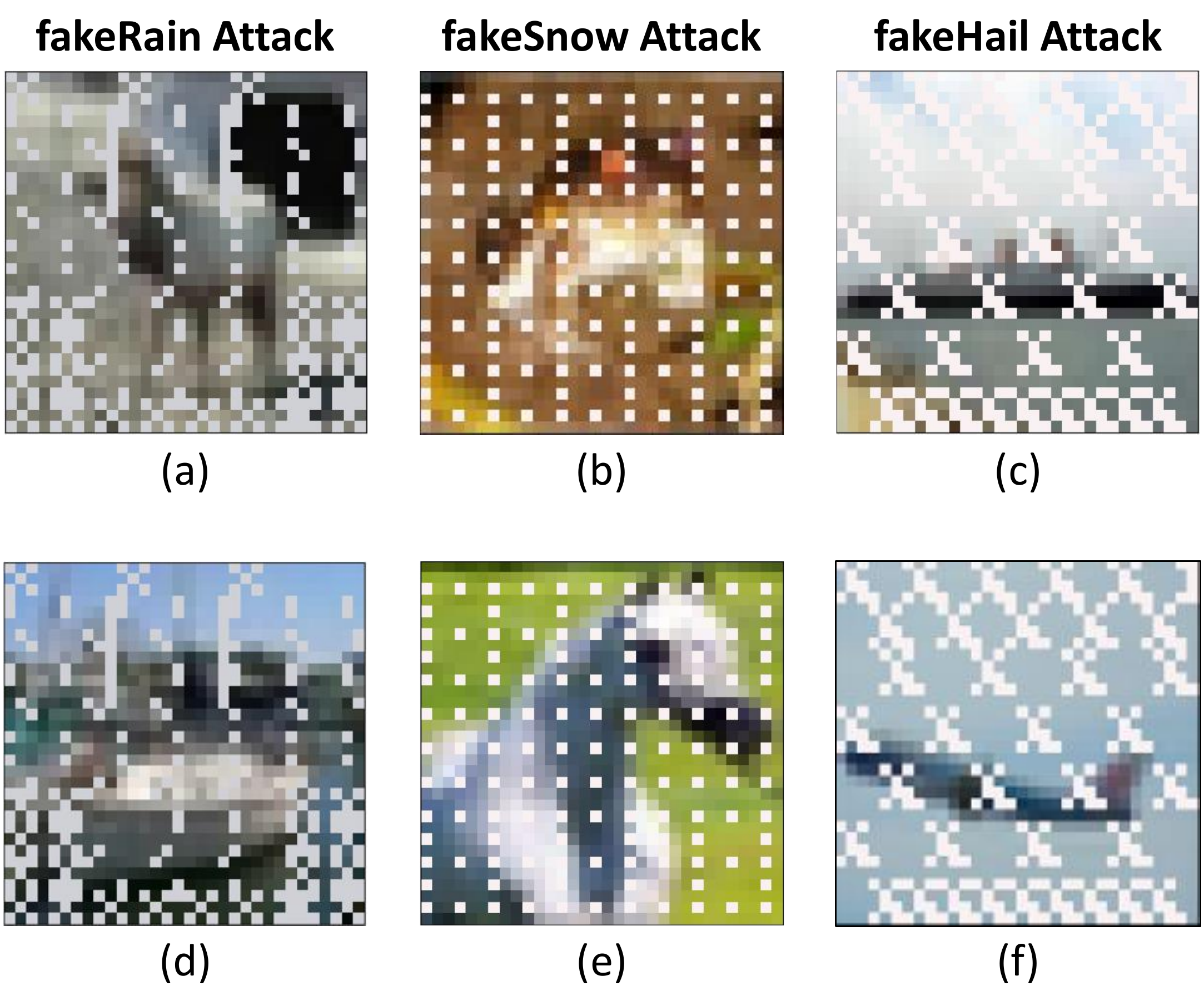}
	\caption{Examples of a few images of the CIFAR-10 dataset on which the \textit{fakeWeather} attacks are applied. (a) and (d): \textit{fakeRain} adversarial examples. (b) and (e): \textit{fakeSnow} adversarial examples. (c) and (f): \textit{fakeHail} adversarial examples.}
	\label{fig:results_examples}
\end{figure}

Table~\ref{tab:ASR_comparison} reports the results for the \textit{fakeRain}, \textit{fakeSnow} and \textit{fakeHail} attacks in terms of Adversarial Success Rate (ASR), which corresponds to the ratio between the misclassified examples and all the tested examples. The results are compared with the state-of-the-art 1-pixel, 3-pixel, and 5-pixel attacks proposed by Su et al.~\cite{Su}. Moreover, Figure~\ref{fig:results_examples} shows a collection of adversarial examples generated with the \textit{fakeWeather} attacks.

\begin{table}[h]
    \centering
    \caption{Evaluation of the Adversarial Success Rate (ASR) for the the LeNet, the ResNet, and the CapsNet on the CIFAR-10 dataset. Our proposed \textit{fakeWeather} attacks have been compared to the 1-pixel, 3-pixel, and 5-pixel attacks~\cite{Su}.}
    \begin{tabular}{cccc}
        \toprule
        \textbf{ASR on Attack} & \textbf{LeNet} & \textbf{ResNet} & \textbf{CapsNet}\\ 
        \midrule
        1-pixel~\cite{Su} & 63\% & 34\% & 19\%\\
        \midrule
        3-pixel~\cite{Su} & 92\% & 79\% & 39\%\\
        \midrule
        5-pixel~\cite{Su} & 93\% & 79\% & 36\%\\
        \midrule
        \textbf{fakeRain (ours)} & 72\% & 67\% & 36\%\\
        \midrule
        \textbf{fakeSnow (ours)} & 75.5\% & 79.5\% & 30\%\\
        \midrule
        \textbf{fakeHail (ours)} & 82.5\% & 78.5\% & 63\%\\
        \bottomrule
    \end{tabular}
    \label{tab:ASR_comparison}
\end{table}

\vspace*{3pt}
\noindent
\textbf{\textit{fakeRain} Evaluation}

The \textit{fakeRain} attack is successful for the LeNet and the ResNet, since their ASRs are $72\%$ and $67\%$, respectively. The ResNet results slightly more robust than the LeNet, due to its deeper structure. The ASR falls to $36\%$ for the CapsNet, since its architecture that groups the neurons into capsules, along with the dynamic routing, helps to better encode the spatial relations between features of the images. The example in \mbox{Figure~\ref{fig:results_examples}a} shows the image of a deer on which the \textit{fakeRain} mask is applied. All the three DNN models erroneously classify it as a ``bird'', while its clean version is correctly classified as a ``deer''. Similarly, the image in \mbox{Figure~\ref{fig:results_examples}d} is incorrectly classified as a ``truck'' by the LeNet and the ResNet, while its clean version is correctly classified as a ``ship''. However, the CapsNet still classifies this adversarial example as a ``ship''.


\vspace*{3pt}
\noindent
\textbf{\textit{fakeSnow} Evaluation}

For the \textit{fakeSnow} attack, the relations between the ASRs of the three DNN models are similar to the observations made for the \textit{fakeRain} attack, in which the CapsNet is more robust than the other CNNs. However, the ASR results are higher for the ResNet, compared to the LeNet. The example in \mbox{Figure~\ref{fig:results_examples}b} showing a frog with the \textit{fakeSnow} mask is correctly classified by the CapsNet, while it is incorrectly classified as a ``cat'' by the ResNet and as a ``truck'' by the LeNet. Its clean version is correctly classified as a ``frog'' by all the DNNs. The horse in \mbox{Figure~\ref{fig:results_examples}e} is correctly classified by the ResNet and the CapsNet, while the LeNet classifies it as a ``deer''.


\vspace*{3pt}
\noindent
\textbf{\textit{fakeHail} Evaluation}

The ASR relative to the \textit{fakeHail} attack is significantly higher than the previous attacks, in particular for the CapsNet. Due to the relatively large perturbations imposed by the hail patterns (i.e., 8-pixel perturbations), the \textit{fakeHail} mask can break the spatial relations learned by the CapsNet and lead to many misclassified samples. The example in \mbox{Figure~\ref{fig:results_examples}c} represents a ship with the \textit{fakeHail} mask that is incorrectly classified as an ``airplane'' by the LeNet and CapsNet, and as a ``truck'' by the ResNet. The image in \mbox{Figure~\ref{fig:results_examples}f} is incorrectly classified as a ``cat'' by the LeNet, as a ``deer'' by the ResNet, and as a ``frog'' by the CapsNet, despite showing an airplane.




\subsection{Case Studies: Output Probability Variations under \textit{fakeWeather} attacks.}

Towards a more comprehensive evaluation, we analyze the output probability variations when different types of \textit{fakeWeather} attacks are applied to the LeNet, ResNet, and CapsNet models. For reference, the $10$ classes of the \mbox{CIFAR-10} dataset are associated with a digit $0-9$ according to the convention in Table~\ref{tab:CIFAR10_classes}.

\begin{table}[h]
    \centering
    \caption{Class labels for the CIFAR-10 dataset~\cite{CIFAR-10}.}
    \begin{tabular}{cc}
        \toprule
        \textbf{\#} & \textbf{Class}\\ 
        \midrule
        0 & airplane\\
        \midrule
        1 & automobile\\
        \midrule
        2 & bird\\
        \midrule
        3 & cat\\
        \midrule
        4 & deer\\
        \midrule
        5 & dog\\
        \midrule
        6 & frog\\
        \midrule
        7 & horse\\
        \midrule
        8 & ship\\
        \midrule
        9 & truck\\
        \bottomrule
    \end{tabular}
    \label{tab:CIFAR10_classes}
\end{table}

\begin{figure*}[h!]
	\centering
	\includegraphics[width=\linewidth]{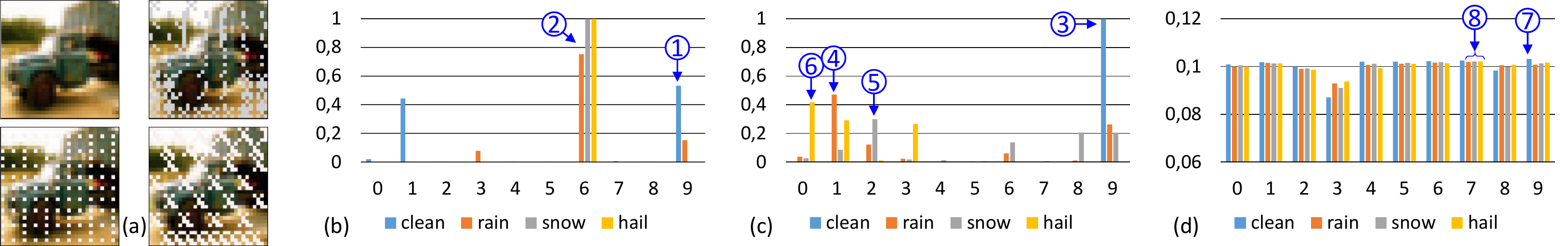}
	\caption{Example showing a ``truck'' to which the \textit{fakeWeather} attacks are applied. (a) Clean image, \textit{fakeRain} image, \textit{fakeSnow} image, and \textit{fakeHail} image. (b) Output probabilities for the LeNet. (c) Output probabilities for the ResNet. (d) Output probabilities for the CapsNet.}
	\vspace*{15pt}
	\label{fig:results_truck}
\end{figure*}

\begin{figure*}[h!]
	\centering
	\includegraphics[width=\linewidth]{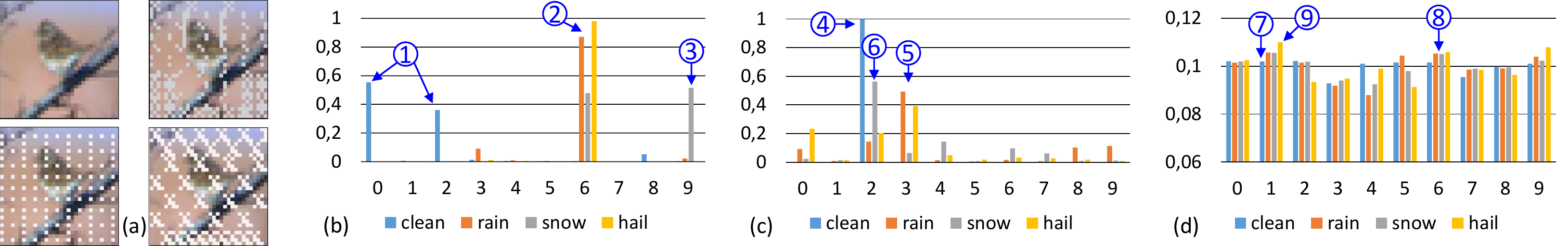}
	\caption{Example showing a ``bird'' to which the \textit{fakeWeather} attacks are applied. (a) Clean image, \textit{fakeRain} image, \textit{fakeSnow} image, and \textit{fakeHail} image. (b) Output probabilities for the LeNet. (c) Output probabilities for the ResNet. (d) Output probabilities for the CapsNet.}
	\label{fig:results_bird}
	\vspace*{5pt}
\end{figure*}

Figure~\ref{fig:results_truck} shows how the image of a ``truck'' of the CIFAR-10 dataset is classified, for different \textit{fakeWeather} attacks and different DNN models. The clean image is correctly classified as the class $9$, i.e.,``truck'' by the LeNet, despite having a relatively low confidence (see pointer~\rpoint{1} in Figure~\ref{fig:results_truck}b). When each of the \textit{fakeWeather} masks is applied, the LeNet predicts the image as a ``frog'' with quite high confidence (see pointer~\rpoint{2} in Figure~\ref{fig:results_truck}b). The probability variations for the ResNet assume a different behavior. While the clean image is correctly classified with high confidence (see pointer~\rpoint{3} in Figure~\ref{fig:results_truck}c), the \textit{fakeWeather} attacks produce different outcomes. With the \textit{fakeRain} mask the image is classified as an ``automobile'' by the ResNet (see pointer~\rpoint{4} in Figure~\ref{fig:results_truck}c), with the \textit{fakeSnow} mask the highest probability belongs to the class ``bird'' (see pointer~\rpoint{5} in Figure~\ref{fig:results_truck}c), and the adversarial \textit{fakeHail} image is classified as an ``airplane'' by the ResNet (see pointer~\rpoint{6} in Figure~\ref{fig:results_truck}c). The output probabilities for the CapsNet, while they are more concentrated in the middle values, i.e., $1/10$, show that the clean image is correctly classified (see pointer~\rpoint{7} in Figure~\ref{fig:results_truck}d), while for all the \textit{fakeWeather} attacks, the highest probability belongs to the class ``horse'' (see pointer~\rpoint{8} in Figure~\ref{fig:results_truck}d).

Figure~\ref{fig:results_bird} shows the output probability variations associated to a ``bird'' image of the CIFAR-10 dataset. The clean image is already incorrectly classified as an ``airplane'' by the LeNet (see pointer~\rpoint{1} in Figure~\ref{fig:results_bird}b). With the \textit{fakeRain} or the \textit{fakeHail} mask, the LeNet classifies the adversarial image as a ``frog'' (see pointer~\rpoint{2} in Figure~\ref{fig:results_bird}b), while the adversarial \textit{fakeSnow} image is classified as a ``truck'' (see pointer~\rpoint{3} in Figure~\ref{fig:results_bird}b). The ResNet correctly classifies the clean image as a ``bird'' with high confidence (see pointer~\rpoint{4} in Figure~\ref{fig:results_bird}c). The \textit{fakeRain} and \textit{fakeHail} adversarial images are classified as a ``cat'' (see pointer~\rpoint{5} in Figure~\ref{fig:results_bird}c), while the \textit{fakeSnow} is unsuccessful, since the image is still correctly classified by the ResNet (see pointer~\rpoint{6} in Figure~\ref{fig:results_bird}c), even though with lower confidence than the clean image. The CapsNet correctly classifies the clean image with very narrow difference w.r.t. the other classes (see pointer~\rpoint{7} in Figure~\ref{fig:results_bird}d). The \textit{fakeRain} and \textit{fakeSnow} attacks produce adversarial images that are classified as a ``frog'' by the CapsNet (see pointer~\rpoint{8} in Figure~\ref{fig:results_bird}d), while the image with the \textit{fakeHail} mask is correctly classified by the CapsNet (see pointer~\rpoint{9} in Figure~\ref{fig:results_bird}d).

\subsection{Results Discussion and Comparison}

To summarize, given the above-discussed results, we can make the following considerations:
\begin{itemize}
    \item All the \textit{fakeWeather} attacks produce a high ASR for the LeNet and ResNet ($ASR > 65 \%$).
    \item The \textit{fakeHail} attack is the strongest, since it achieves an ASR equal to $63\%$ for the CapsNet and higher for the other DNNs. 
\end{itemize}

Compared to the methods of Su et al.~\cite{Su}, our \textit{fakeWeather} methods have higher ASR than the 1-pixel attack for every DNN model (see Table~\ref{tab:ASR_comparison}). However, the \mbox{3-pixel} and \mbox{5-pixel} attacks have higher ASR than our methods. Note that the approach used by Su et al. is based on an evolutionary algorithm that requires several queries, while our methodology does not require any query. Yet, the ASR relative to the CapsNet for the \textit{fakeHail} attack is $27\%$ higher than the \mbox{5-pixel} attack.

\subsection{Future Outlooks and Applicability}

From another perspective, our contributions, other than a methodology for generating adversarial attacks in \mbox{real-time} without queries, can be viewed as a data augmentation methodology for generating synthetic samples of weather conditions. We envision the possibility of enlarging the dataset with images that contain \textit{fakeWeather} masks and train DNN-based classifiers more robustly to such atmospheric phenomena, in a similar way as the adversarial training's functionality~\cite{Madry2018PGD}. Since the only information required is the image size, its high scalability makes our \textit{fakeWeather} attack methodology suitable to any vision-based outdoor application.




\section{Conclusion}
In this paper, we presented \textit{fakeWeather} attacks, adversarial attacks for DNNs that emulate the natural weather conditions. Our methodology consists of observing a series of images that capture the effects of such conditions perceived by the camera lens, and modeling a set of patterns to create dedicated \textit{fakeRain}, \textit{fakeSnow}, and \textit{fakeHail} masks as a collection of these patterns. Hence, these sets of perturbations make the adversarial image a plausible input to the DNN. Our proposed attack is conducted in \textit{true \mbox{black-box}} settings, in which the adversary has no access to the DNN model, its parameters, and its output. The evaluation of \textit{fakeWeather} attacks on different DNN models (Convolutional Neural Networks and Capsule Networks) highlights noticeable adversarial success rates.

\section*{Acknowledgment}

This work has been supported in part by the Doctoral College Resilient Embedded Systems, which is run jointly by the TU Wien's Faculty of Informatics and the UAS Technikum Wien. This work was also supported in parts by the NYUAD Center for Interacting Urban Networks (CITIES), funded by Tamkeen under the NYUAD Research Institute Award CG001, Center for CyberSecurity (CCS), funded by Tamkeen under the NYUAD Research Institute Award G1104, and Center for Artificial Intelligence and Robotics (CAIR), funded by Tamkeen under the NYUAD Research Institute Award CG010.

\begin{refsize}
\bibliographystyle{ieeetr}
\bibliography{main.bib}
\end{refsize}

\end{document}